\documentclass[10pt,nocopyrightspace]{ewsn-proc}

\usepackage{graphicx}
\usepackage{balance}
\usepackage{comment}
\usepackage{siunitx}
\usepackage{xfrac}
\usepackage{xcolor}
\usepackage{booktabs}
\usepackage{multirow}
\usepackage[hidelinks]{hyperref}
\usepackage{mathrsfs}

\graphicspath{{images}}
\DeclareGraphicsExtensions{.png}

\makeatletter
\newcommand{\todo}[1]{\noindent\textit{\color{red}\textbf{TODO}~#1}\@latex@warning{TODO: #1}} 
\makeatother

\DeclareSIUnit\nothing{\relax}
\DeclareSIUnit\mac{MAC}

\definecolor{somegray}{rgb}{0.5, 0.5, 0.5}
\newcommand{\darkgrayed}[1]{\textcolor{somegray}{#1}}
\makeatletter
\newcommand*\titleheader[1]{\gdef\@titleheader{#1}}
\AtBeginDocument{%
  \let\st@red@title\@title
  \def\@title{%
    \vskip-2.0em
    \bgroup\normalfont\large\centering\@titleheader\par\egroup
    \vskip1.0em\st@red@title}
}
\makeatother

\titleheader{\darkgrayed{This paper has been accepted for publication in the EWSN 2023 conference\\\copyright 2023 ACM.}}

\numberofauthors{3}

\author{
\alignauthor Elia Cereda \\
    \affaddr{IDSIA, USI-SUPSI, Switzerland}\\
    \affaddr{~}\\
    \email{elia.cereda@idsia.ch}
\alignauthor Alessandro Giusti \\
    \affaddr{IDSIA, USI-SUPSI, Switzerland}\\
    \affaddr{~}\\
    \email{alessandro.giusti@idsia.ch}
\alignauthor Daniele Palossi \\
    \affaddr{IDSIA, USI-SUPSI, Switzerland}\\
    \affaddr{IIS, ETH Zurich, Switzerland}\\
    \email{daniele.palossi@idsia.ch}
}

\title{Secure Deep Learning-based Distributed Intelligence on Pocket-sized Drones}

\begin{document}

\maketitle

\begin{abstract}
Palm-sized nano-drones are an appealing class of edge nodes, but their limited computational resources prevent running large deep-learning models onboard. 
Adopting an edge-fog computational paradigm, we can offload part of the computation to the fog; however, this poses security concerns if the fog node, or the communication link, can not be trusted. 
To tackle this concern, we propose a novel distributed edge-fog execution scheme that validates fog computation by redundantly executing a random subnetwork aboard our nano-drone.
Compared to a State-of-the-Art visual pose estimation network that entirely runs onboard, a larger network executed in a distributed way improves the $R^2$ score by +0.19; in case of attack, our approach detects it within \SI{2}{\second} with 95\% probability.
\end{abstract}

%
%

%

\section*{Supplementary material}
In-field experiments and system demonstration video: \href{https://youtu.be/QwTiigAs4cA}{https://youtu.be/QwTiigAs4cA}.

\section{Introduction} \label{sec:intro}

With an increasing number of Internet-of-Things-capable (IoT) end devices, from tiny headphones to autonomous cars, the \textit{edge-fog paradigm} permeates almost any civil and industrial application~\cite{khanna2020internet}.
This paradigm exploits distributed computation with a resource-limited edge device close to the source of data, combined with a more capable remote fog node able to overcome the memory and computational limits of the former node.
As the edge node, this work considers a novel and appealing cyber-physical system: a pocket-sized quadrotor or nano-drone.
Thanks to their sub-\SI{10}{\centi\meter} diameter, nano-drones have the potential to unlock unprecedented application scenarios, such as the exploration of narrow, cluttered or GPS-denied environments, safe human-robot interaction, and intelligent ubiquitous IoT sensing.
Additionally, these platforms are much less expensive than larger ones due to simplified electronics and mechanics.

However, with their small form factor comes their main limitation: they can only host ultra-constrained processors and sensors, i.e., simple ultra-low power microcontroller units (MCUs) with a sub-\SI{100}{\milli\watt} power envelope.
This limitation poses two challenges; on the one hand, they have extremely limited computational power and memory to run aboard complex real-time algorithms.
To partially overcome this issue, nano-drones often adopt navigation engines based on small convolutional neural networks (CNNs)~\cite{pulp-frontnet, 9869931}, due to their compelling trade-off between accuracy and computational cost, compared to geometrically precise computer vision-based~\cite{6595932} and predictive methods~\cite{9654841}.

On the other hand, the security features nano-drones can offer are still quite rudimentary and rarely addressed by State-of-the-Art (SoA) solutions~\cite{9399464}.
An edge-fog scheme introduces two significant vulnerabilities in the remote fog node and the communication channel: compromising either of them, e.g., with fog malware, man-in-the-middle, data infiltration, or data-in-transit attacks, allows an attacker to send malicious commands to the edge and take control of its behavior.
Our work tries to mitigate both vulnerabilities.

\begin{figure}
\centering
\includegraphics[width=\columnwidth]{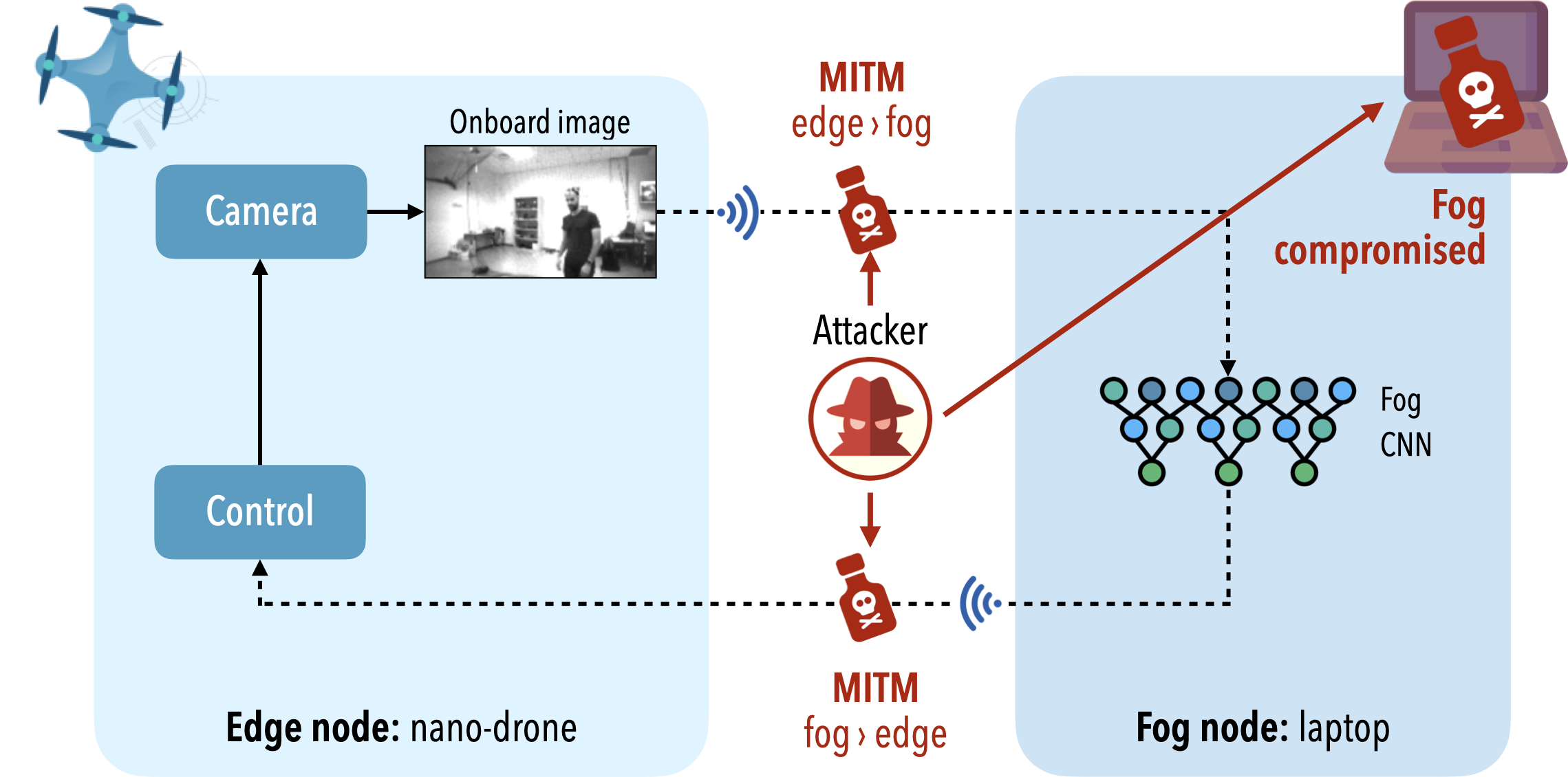}
\caption{The threat model of our edge-fog use case.}
\label{fig:threat_model}
\end{figure}

We address the human pose estimation task and embody the edge-fog paradigm with a nano-drone (edge) and a WiFi-connected commodity laptop (fog), which acts as a powerful remote brain to boost the perception capabilities of the edge node, as depicted in Figure~\ref{fig:threat_model}.
In our case, the edge-fog execution scheme enables execution of more complex CNNs, otherwise unaffordable aboard the nano-drone within its real-time constraints.
We design a three-stage distributed MobileNetV2-based~\cite{mobilenetv2} CNN to perform our vision-based task.
The first compression stage transforms a large input image into a smaller tensor; a central backbone performs the most computationally-intense processing; a final reduction stage produces the output.
Then, we design the central backbone with a multi-branch structure, where multiple tensors are computed independently and in parallel before being reduced in the final stage.
We execute the first and last stages on the nano-drone while we offload the heavy multi-branch backbone to the fog node.

Our strategy enables an additional level of security between edge and fog nodes, which can be stacked on top of traditional encryption mechanisms if affordable by the edge MCU.
While the fog executes all branches for every input image, the edge also executes one random branch at a time and checks its resulting tensor against the one received from the fog, embodying a probabilistic security mechanism.
If the check is successful, the edge uses all the received tensors (one per branch) to run the final reduction.
Otherwise, it falls back to a small field-proofed CNN able to run in real-time aboard a nano-drone, i.e., the PULP-Frontnet~\cite{pulp-frontnet}.
The PULP-Frontnet acts as a backup solution, taking control when either the information received by the fog is compromised or when the communication channel is disrupted/unavailable (e.g., jamming).
This shallow CNN trades the regression accuracy w.r.t. the more accurate remote multi-branch CNN for lower complexity, making it our ``last resort'' to complete or gracefully abort the mission.

We demonstrate our general approach by showing the perceptive benefit of a multi-branch MobileNetV2-based CNN, which achieves a mean $+0.19$ $R^2$ score increase compared to the PULP-Frontnet baseline.
Then, we introduce a simple model accounting for computation and communication overheads to identify the optimal points to split our CNN between compression, backbone, and reduction stages to maximize the overall throughput up to $\sim$\SI{8}{frame/\second}.
Finally, we deploy our distributed system in the field, assessing its security functionalities (see in-field video) and closed-loop performance with a 31.3\% (horizontal) and 34.6\% (angular) reduction of the control error compared to the baseline.
\section{Related work} \label{sec:related}

Autonomous navigation algorithms executed aboard nano-robotics platforms are subject to severe computational and memory constraints.
SoA CNNs deployed on nano-drones exploit novel parallel ultra-low-power systems-on-chip (SoCs)~\cite{nanoflownet,pulp-dronet,pulp-frontnet}, but they can only execute much simpler real-time workloads compared to those running on larger-scale UAVs.
Therefore, limiting the complexity of the achievable tasks and the quality of the obtained results.
Among these works, PULP-Frontnet~\cite{pulp-frontnet} is a monocular CNN for the human pose estimation task with a computational cost of \SI{14.7}{\mega\mac} (millions of multiply-accumulate operations). 
It achieves an inference rate of \SI{48}{frame/\second} while consuming \SI{96}{\milli\watt}, aboard a Crazyflie 2.1 nano-drone.

By comparison, common visual CNN architectures~\cite{scaramuzza_alphapilot,deeppilot_2020,jung2018perception,dronecap} deployed on larger drones have orders of magnitude higher computation workloads.
For example, Foehn et al.~\cite{scaramuzza_alphapilot} tackle the task of gate pose estimation in an autonomous drone race using a \SI{1930}{\mega\mac} per inference CNN running on an Nvidia Jetson Xavier board ($\sim$\SI{20}{\watt}).
These systems either depend on more powerful on-board processors than available on nano-drones~\cite{scaramuzza_alphapilot,jung2018perception} or offload the computation to a remote node and thus suffer from the security vulnerabilities we address in this work~\cite{deeppilot_2020,dronecap}. 

Several ad-hoc neural network architectures have been proposed for efficient execution while retaining SoA task performance, such as MobileNetv2~\cite{mobilenetv2} or EfficientNetv2~\cite{efficientnetv2}.
We select MobileNetv2, i.e., one of the most widespread CNN for embedded devices, as the base architecture for our work.
Given its design for smartphone-class devices with \SI{90}{\mega\mac} per inference, we leverage a fog node to achieve real-time throughput and SoA regression performance on our task.
Additionally, despite the specific architecture used in our use case, we present a general approach that can be applied to other CNNs and for different tasks.

We extend MobileNetv2 with a distributed execution scheme based on multiple independent branches to achieve the desired security properties.
Multi-branch architectures have been adopted in several recent works for various reasons, such as \textit{i}) reducing the computational cost by dynamically executing only subsets of the branches~\cite{hydranet,branchynet}, \textit{ii}) to estimate uncertainty in network predictions~\cite{deepsubensembles}; \textit{iii}) to prevent adversarial attacks~\cite{multibranchadversarial}.
Crucially, these works all focus on models running on a single device, while our work is the first to employ a multi-branch architecture to detect attacks on a distributed edge-fog system, where an agile nano-drone embodies the edge node.
\section{System design}

\subsection{Use case}
In the envisioned scenario, the nano-drone (edge node) is tasked with a vision-based perception task, i.e., human pose estimation.
A more capable fog node, represented by a resource-unconstrained remote laptop, provides additional computational power to the nano-drone to increase its execution performance.
Edge and fog nodes share a bi-directional communication channel, i.e., WiFi, on which they can continuously stream data.
We select two different deep-learning models to perform the pose estimation task.
The former is a lightweight (\SI{300}{\kilo\nothing} parameters) convolutional neural network (CNN) called PULP-Frontnet~\cite{pulp-frontnet} and previously demonstrated running in real-time aboard nano-drones.
The latter is a more memory-/computationally-demanding (\SI{1.8}{\mega\nothing} parameters) but also more accurate CNN, based on the MobileNetV2 model~\cite{mobilenetv2}. 
The MobileNetV2-based workload is distributed between edge and fog, where most of it is executed by the more capable remote fog, leaving a smaller fraction of computation on the memory/processor-limited edge.
The lightweight PULP-Frontnet is executed entirely on the edge, only if the communication channel or the remote computation becomes compromised, as detailed in the following.

\subsection{CNN design} \label{sec:CNN_design}
\begin{figure}
\centering
\includegraphics[width=\columnwidth]{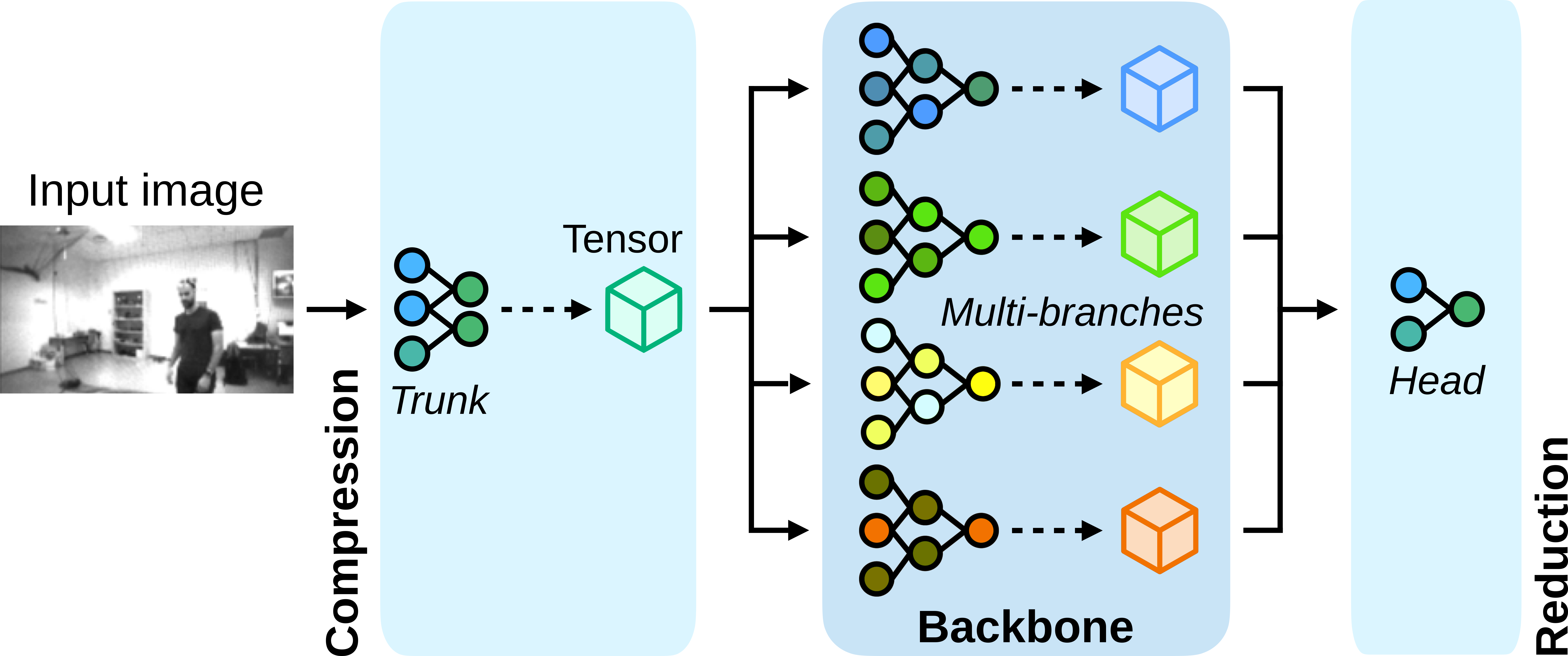}
\caption{Vision-based CNN's three main stages.}
\label{fig:CNN_split}
\end{figure}

The traditional execution scheme of many vision-based CNNs can be mapped into three main computational stages, as shown in Figure~\ref{fig:CNN_split}.
\textit{i}) an initial compression stage, where a high-resolution input image is compressed to a smaller-size tensor; \textit{ii}) the execution of a central computationally-demanding backbone; and \textit{iii}) a final reduction stage, e.g., a fully connected layer, producing the final output.
Under this consideration, we design our MobileNetV2-based CNN as a multi-branch model, where the central backbone is split into $\mathcal{N}$ independent branches.
Each branch is fed with the same input tensor, i.e., the output of the initial compression stage -- called \textit{trunk} -- and produces its output tensor.
The last reduction stage, called \textit{head}, takes as input the concatenation of all tensors resulting from the multi-branch backbone and produces the final output for the robot controller.

We apply this general approach to the task of human pose estimation.
Our CNN takes as input one gray-scale $160\times96$ px image and estimates the pose of the human subject relative to the drone.
The pose components are represented as four independent regression outputs, which correspond to the three Cartesian coordinates of the subject’s position in 3D space $(x, y, z)$ w.r.t. the drone’s horizontal frame and the subject’s relative orientation w.r.t. the drone’s yaw, $\phi$.
Finally, we employ this relative pose ($x, y, z, \phi$) to perform the ``follow-me'' application aboard our nano-drone.

\subsection{Threat model}
The success of our mission depends on \textit{i}) correct execution of both edge and fog computation, and \textit{ii}) reliable and unaltered data exchange between edge and fog nodes.
We focus on the two vulnerabilities shown in Figure~\ref{fig:threat_model}, in which a malicious actor either
compromises the entire fog node or the communication channel, e.g., man-in-the-middle, data infiltration, and data-in-transit hacked attacks.
Either scenario would allow the attacker to poison the information transmitted by the fog node and thus take control of the nano-drone.
To mitigate this threat, we introduce a probabilistic security mechanism that relies on redundant execution of a small amount of computation on the edge node to assess the trustworthiness of the data received from the fog.

Our approach is orthogonal to computationally-demanding traditional countermeasures, such as communication encryption (not always affordable on MCU-limited edge nodes), providing a \textit{defense-in-depth} strategy.
Finally, the last attack we consider is complete loss of the edge-fog communication channel (e.g., radio jamming).
In this case, we resort to safe execution of the PULP-Frontnet CNN entirely aboard our nano-drone, degrading pose estimation accuracy but without jeopardizing the mission.

\subsection{Proposed strategy}
\label{sec:proposed_strategy}

\begin{figure}
\centering
\includegraphics[width=\columnwidth]{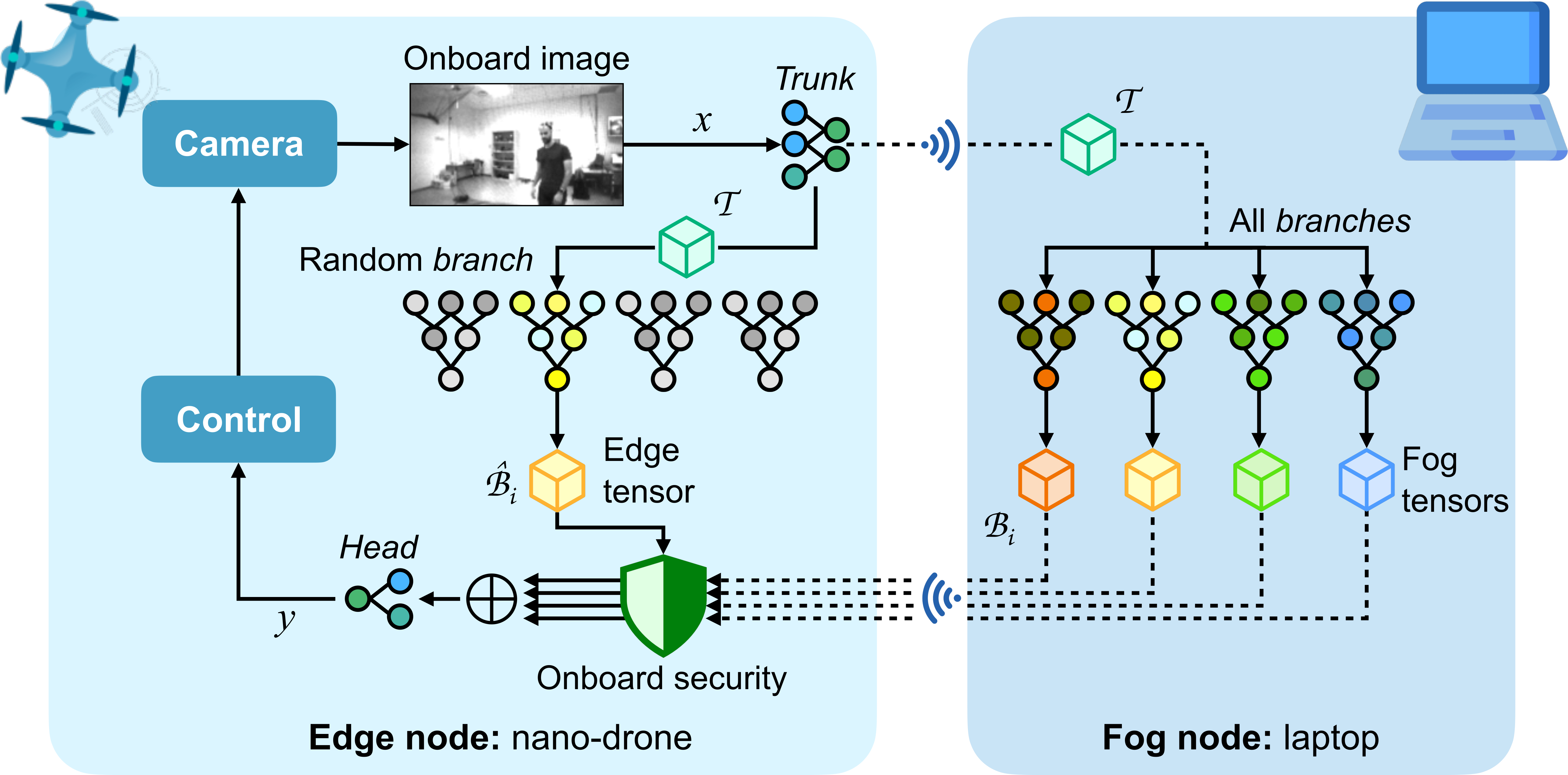}
\caption{Our distributed probabilistic security strategy.}
\label{fig:approach}
\end{figure}

We propose a strategy based on partial computation redundancy to enable an onboard probabilistic security mechanism while increasing the edge inference throughput thanks to the computationally-capable fog.
Figure~\ref{fig:approach} shows the proposed system.
The execution starts from the edge node running the first compression stage (\textit{trunk} CNN) on the input image ($x$) and producing a compressed tensor $\mathcal{T}$ to reduce the streaming data between edge and fog nodes:
\begin{equation}
    \mathcal{T} = \mathit{trunk}(x)
\end{equation}
$\mathcal{T}$ is then forwarded via WiFi to the fog and used locally on the edge.
The fog node executes the entire multi-branch backbone for each input tensor $\mathcal{T}$.
The CNN running on the fog is the backbone (central part) of a MobileNetV2 CNN composed of $\mathcal{N}$ independent branches, where $\mathcal{N}$ is a hyperparameter of the model.
Each branch receives the same input $\mathcal{T}$ and produces one unique feature tensor $\mathcal{B}_i$:
\begin{equation}
    \mathcal{B}_i = \mathit{branch}_i(\mathcal{T}) \qquad \text{for } i = 1 ... \mathcal{N}
\end{equation}
Feature tensors from all branches are then forwarded via WiFi to the edge.

On the edge node, only one random branch $j$ is executed per input image to minimize the computation burden. 
A cryptographic random number generator ensures that an attacker cannot predict the sequence of verified branches.
The selected branch is fed with $\mathcal{T}$ and produces $\hat{\mathcal{B}_j}$, which is compared against the fog's corresponding computation $\mathcal{B}_j$.
If $\mathcal{B}_j$ is bit-by-bit identical to $\hat{\mathcal{B}_j}$, then all the tensors received from the fog are considered trustable and concatenated ($\bigoplus$ operator) before being reduced in the final stage:
\begin{equation}
    y = \mathit{head}(\bigoplus_i \mathcal{B}_i)
\end{equation}
where the \textit{head} CNN computes the final output $y$ on the edge.

Under this system design, the attacker maximizes their chances of avoiding detection by tampering with only a single feature tensor $\mathcal{B}_k$.
This translates to a probability $\sfrac{1}{\mathcal{N}}$ of detecting an attack in any given frame (i.e., the probability that the edge node executes the same branch corrupted by the attacker $j = k$).
Nevertheless, our detection scheme is so lightweight that it can run continuously as part of a robot's closed-loop controller.
Therefore, the attacker must continue tampering over time to keep control of the nano-drone, but our attack detection probability converges to 1.0 as more frames are verified.

\subsection{Deployment}

\textbf{Robotic platform.}
Our edge node is embodied by a commercial off-the-shelf (COTS) Crazyflie~2.1 nano-quadrotor, an open-source \SI{27}{\gram} palm-sized nano-drone produced by Bitcraze.
An STM32 MCU performs low-level flight control, while an AI-deck COTS expansion board provides high-level vision-based perception with an additional GWT GAP8 SoC.
The GAP8 is a multi-core processor featuring 9 RISC-V-based cores, split between an eight-core \textit{cluster} optimized for parallel computation of compute-intense workloads and a single-core \textit{fabric controller} that manages interfaces with external peripherals and on-chip memories.
The on-chip memory hierarchy is organized into two levels: \SI{64}{\kilo\byte} low-latency L1 memory and \SI{512}{\kilo\byte} L2 memory.
This expansion board also provides off-chip memories (\SI{8}{\mega\byte} DRAM and \SI{64}{\mega\byte} Flash), a monochrome QVGA Himax HM01B0 camera, and an ESP32-based Ublox NINA-W102 WiFi module.
The lack of hardware floating-point units and data cache memories on the GAP8 SoC dictate, respectively, integer-quantized arithmetic and explicit data management between memories.
We leverage an ad-hoc CNN deployment pipeline to generate C code that addresses these concerns~\cite{pulp-frontnet}.
The fog node also runs its workload in fixed-point arithmetic to simplify bit-by-bit tensor comparisons.

\textbf{Multi-branch architecture.}

\begin{figure}
\centering
\includegraphics[width=\columnwidth]{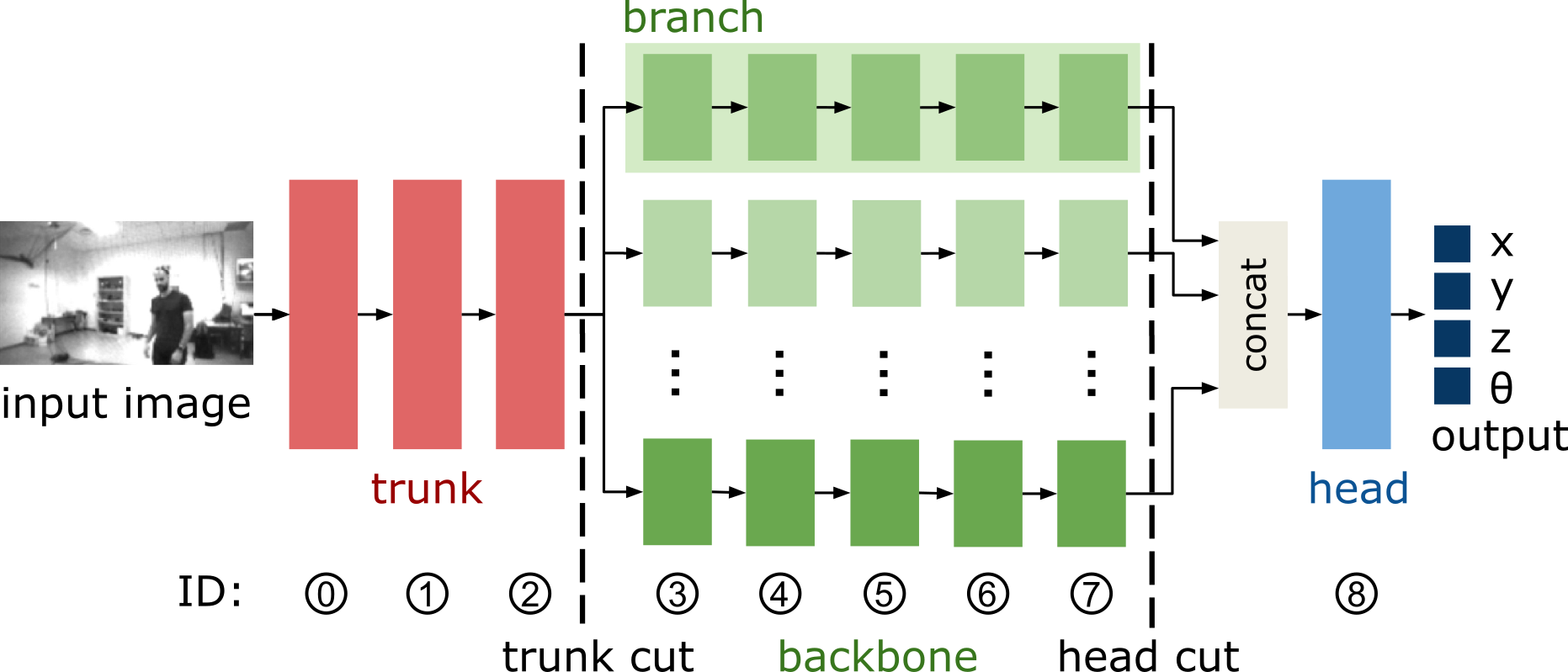}
\caption{Multi-branch MobileNetV2 CNN architecture.}
\label{fig:multi_branch_arch}
\end{figure}

We design our CNN starting from the widely adopted MobileNetV2~\cite{mobilenetv2}, which requires \SI{89}{\mega\mac} operations per inference.
First, we define a coarse split of the three CNN segments we introduced in Section~\ref{sec:CNN_design}: the trunk, a multi-branch backbone, and the final head, see Figure~\ref{fig:multi_branch_arch}.
Branching the execution from the beginning of the CNN, which means discarding the trunk, would lead to poor regression performance and a waste of resources by neglecting an initial trainable path shared by all branches~\cite{2014visualizecnn}.
Therefore, initially, we define as the \textit{cutting point} between the trunk and backbone the first layer having an output tensor smaller than the input image ($\sim$\SI{15}{\kilo\byte}), which results in a trunk of \SI{25.7}{\mega\mac} operations.
The remaining \SI{63.3}{\mega\mac} are equally distributed among all $\mathcal{N}$ branches, splitting the number of output channels.
For example, in the case of $\mathcal{N}=8$, a layer producing an output tensor of 32 channels in the original CNN is split into 8 layers, each outputting a tensor of 4 channels.

\begin{figure}
\centering
\includegraphics[width=0.95\columnwidth]{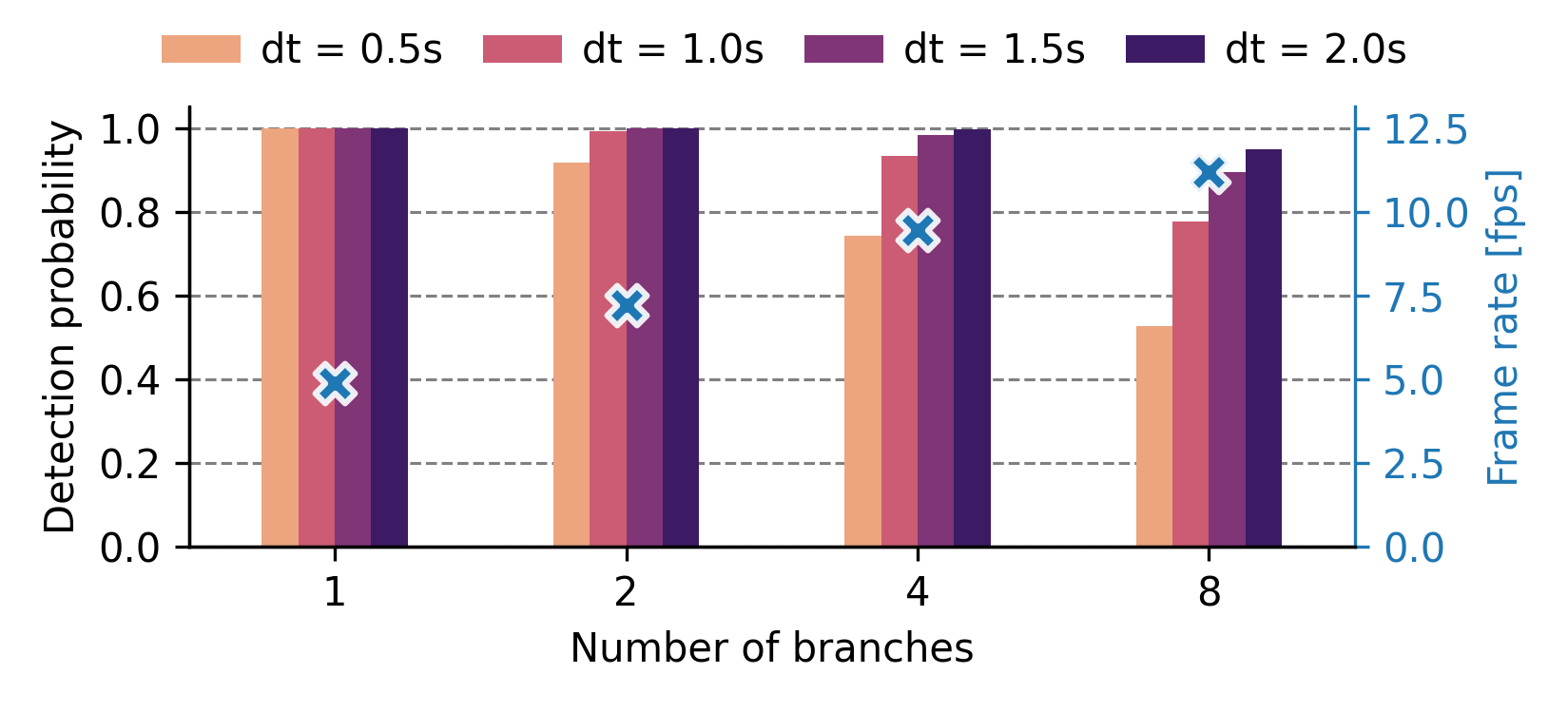}
\caption{Detection probability vs. inference frame-rate.}
\label{fig:attack_detect_dt}
\end{figure}

Figure~\ref{fig:attack_detect_dt} shows four configurations in the number of backbone's branches, i.e., 1, 2, 4, and 8.
For each configuration, we report on the primary y-axis the probability of detecting an attack within a $\mathit{dt}$ time of 0.5, 1, 1.5, and \SI{2}{\second}; the higher the $\mathit{dt}$, the higher the probability of converging to 1.
On the secondary y-axis, we show the achievable frame rate for each configuration; more branches result in less computation on the edge node and lower detection probability. 
The real-time constraint posed by our cyber-physical system suggests selecting a configuration with a frame rate as higher as possible, with a lower bound of $\approx$\SI{10}{frame/\second}~\cite{pulp-frontnet}.
Therefore, we use the configuration $\mathcal{N}=8$ for our experiments, which also achieves a detection probability of 95\% with a $\mathit{dt}$=\SI{2}{\second}.

\begin{figure*}[t]
\centering
\includegraphics[width=\textwidth]{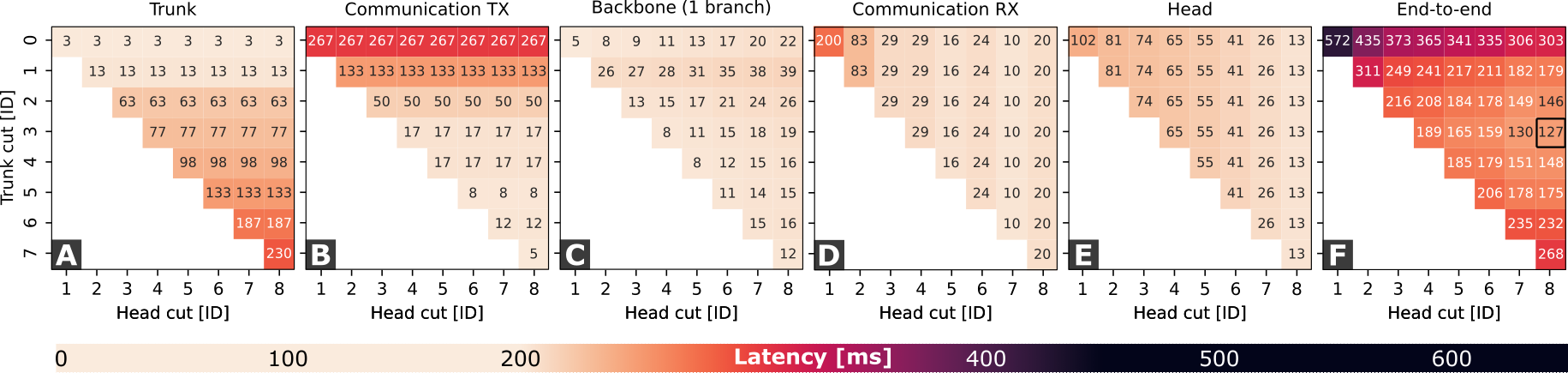}
\caption{Latency induced by each component, for every combination of cut points.}
\label{fig:pipeline_latency}
\end{figure*}

Finally, in Figure~\ref{fig:pipeline_latency}, we thoroughly analyze how different cutting points affect the edge execution time and TX/RX communication latencies between edge and fog, for which we assume a negligible execution time having multiple orders of magnitude more capable compute unit.
Layers in the MobileNetV2 architecture are grouped in \textit{inverted residual blocks}, inside which tensor sizes follow an expansion-compression pattern.
For this reason, we force cutting points to be placed at block boundaries, which we show in Figure~\ref{fig:multi_branch_arch} with 9 IDs.
Earlier cutting points in the trunk, y-axis in Figure~\ref{fig:pipeline_latency}, minimize its execution time (A) but inflates the TX communication latency (B) as more data are streamed to the fog.
Similarly, later cutting points of the head, x-axis in Figure~\ref{fig:pipeline_latency}, increase the execution time of the trunk (A), the backbone's one branch (C), but reduce the latencies for both RX communication (D) and the head execution (E).
End-to-end latency (F) considers that the edge node executes the backbone branch in parallel with communication:
\begin{equation}
    \textit{end-to-end} = \textit{trunk} + \mathrm{max}(\textit{backbone}, \textit{TX} + \textit{RX}) + \textit{head}
\label{eq:e2e}
\end{equation}
The resulting optimal configuration is trunk cutting point ID = 3 and head cutting point ID = 8.
Finally, we deploy and profile this optimal model on the GAP8 SoC, obtaining a $\pm5\%$ execution time compared to the estimates in Figure~\ref{fig:pipeline_latency}.
\section{Results}

We report three experiments.
The first shows the improved regression performance of our proposed multi-branch model.
The second shows the resulting improvement of in-field robot behavior.
The third demonstrates the attack detection capabilities of the proposed security scheme.

\subsection{Regression performance}
We use the dataset presented in~\cite{ceredahandling} for our training and regression performance analysis.
This dataset contains images acquired with the same robotic platform employed in our work and labels of the absolute pose of human subjects and a nano-drone.
Data are collected from two indoor laboratories equipped with a motion capture system (mocap), accounting for 12k images collected with 17 distinct human subjects (age, height, etc.).
Three subjects (4.7k samples) form our test set, while the remaining 14 subjects (7.3k) are used for training (90\%) and validation (10\%).
Starting from models pre-trained on ImageNet, we train for 100 epochs with the Adam optimizer and learning rate $10^{-3}$ to minimize the L1 loss of the relative pose.

Regression performance represents the ability of the model to estimate the subject's relative pose accurately.
We quantify it through the $R^2$ score (or \emph{coefficient of determination}) computed for each output variable.
 $R^2$ is a standard adimensional metric representing the fraction of variance in the target variable explained by the model. 
An ideal model that perfectly estimates the variable yields $R^2 = 1$; a trivial model that always returns the test set average yields $R^2 = 0.0$.

\begin{figure*}[t]
\centering
\includegraphics[width=0.9\textwidth]{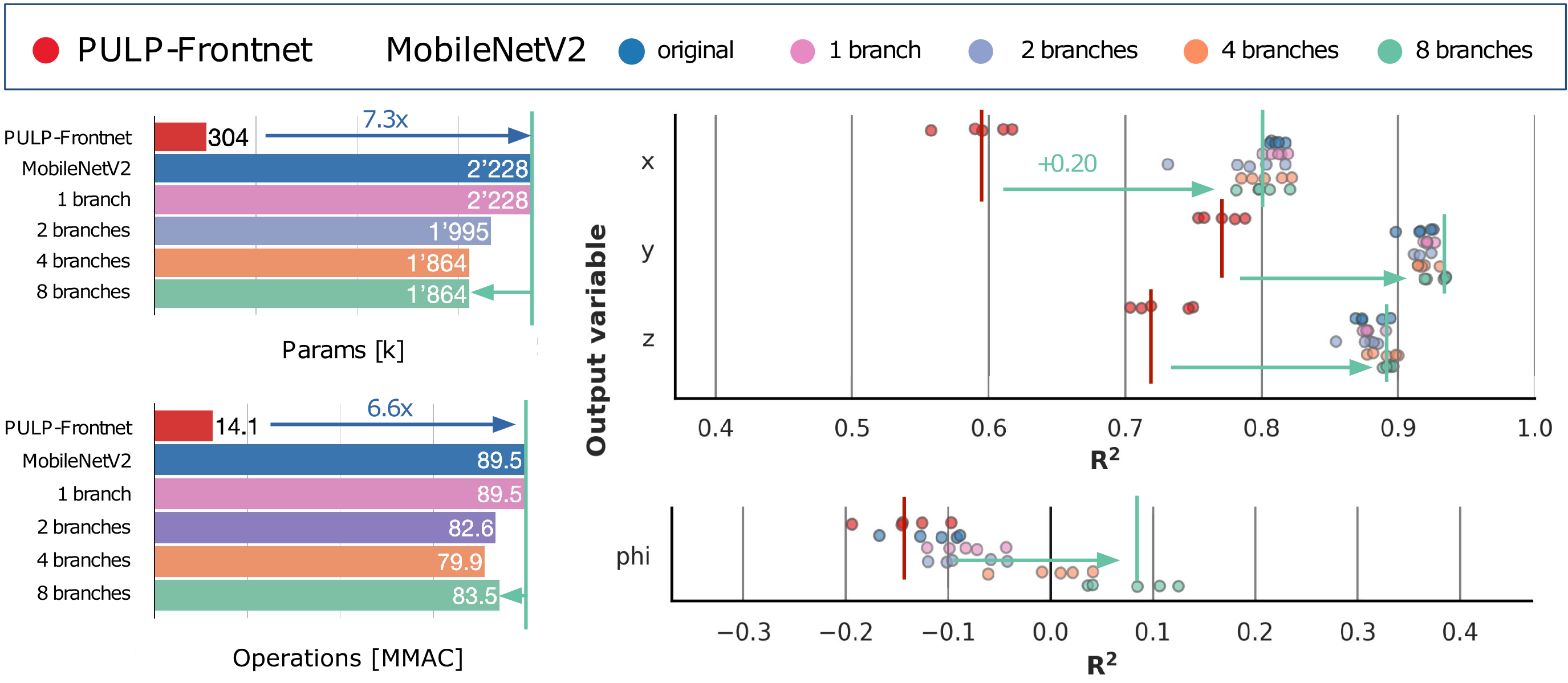}
\caption{CNN parameters (top left), operations (bottom left), and regression performance (right).}
\label{fig:results_r2}
\end{figure*}

\begin{figure}[t]
\centering
\includegraphics[width=0.95\columnwidth]{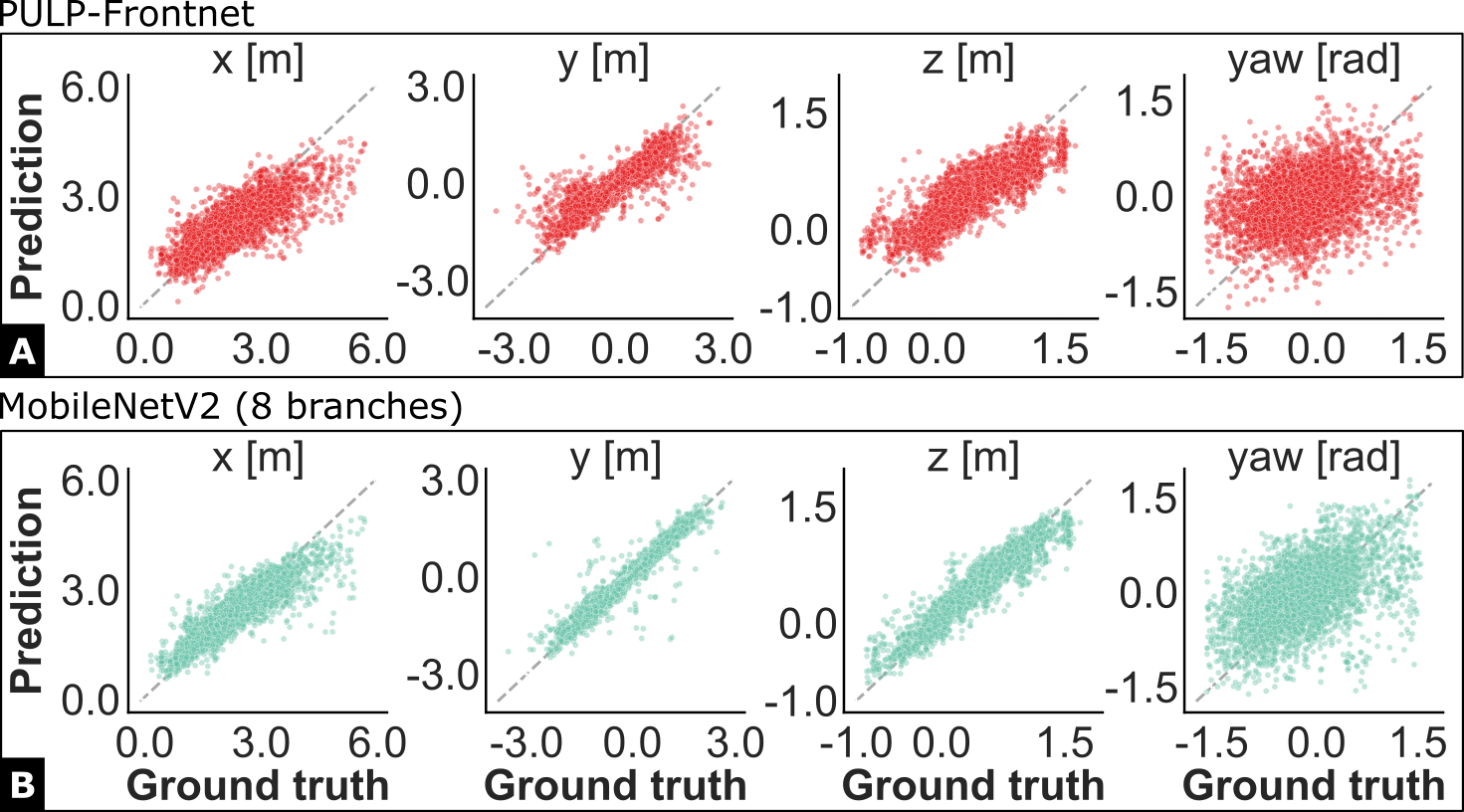}
\caption{Scatter plots of predictions vs. ground truth.}
\label{fig:results_scatters}
\end{figure}

Figure~\ref{fig:results_r2} compares the proposed fog+edge multi-branch models against the SoA edge-only PULP-Frontnet and the fog-only original MobileNetV2.
Each model is trained five times with randomly-initialized parameters.
The original MobileNetV2 shows a $\sim7\times$ increase in network parameters and computation workload, which pays off with an increase of more than $+0.20$ in median $R^2$ score compared to PULP-Frontnet, justifying the use of a fog node and larger models.
Compared to the original MobileNetV2, multi-branch models yield similar (or slightly improved) computation workloads and $R^2$ scores on all output variables. 
Thus, introducing the multi-branch architecture enables our security application without impacting other aspects of the system.
Figure~\ref{fig:results_scatters} compares individual predictions of the 8-branch MobileNetV2 against PULP-Frontnet, showing a significantly lower regression error on all four pose components.

\subsection{Control performance}
This experiment evaluates our model's tracking accuracy when used in a fully-autonomous in-field closed-loop control system.
We reproduce the testing setup of our baseline~\cite{pulp-frontnet}, where the human subject, never seen by the models during training, moves along a predefined path of increasing difficulty.
At the same time, the autonomous drone is tasked to stay in front of the subject at a constant distance of \SI{1.5}{\meter}, therefore behaving in the ``follow-me'' application.
We compare two models: the SoA PULP-Frontnet baseline, running entirely on the edge node, and the 8-branch MobileNetV2 model, running remotely on the fog node, which is also affected by the end-to-end communication latency.
While a mocap system tracks both subject and drone poses, we perform three flights per model ($3 \times 2 = 6$ flights) plus one additional flight in which the controller relies on the subject's ground-truth pose (as measured by the mocap), representing the behavior with a perfect predictor.

Table~\ref{tab:control_error} measures the resulting control performance through two error terms: $e_{xy}$, the drone's mean horizontal distance from its desired position (i.e., \SI{1.5}{\meter} in front of the subject), and 
$e_{\theta}$, the drone's mean absolute angular error from its desired orientation (i.e., looking directly at the subject).
The mocap-based flight represents the control error lower bound achievable with perfect predictions.
We observe that the 8-branch MobileNetV2 model significantly improves on the baseline, reducing both errors by -30\%.
The supplementary video also shows these experiments, highlighting the improved control accuracy.

\begin{table}
\centering
\caption{Control error.}
\label{tab:control_error}
\resizebox{\columnwidth}{!}{
\renewcommand{\arraystretch}{1.25}
    \begin{tabular}{llccc}
    \toprule
    & & \multirow{2}[2]{*}{PULP-Frontnet~\cite{pulp-frontnet}} & \multirow{2}[2]{*}{\shortstack[c]{MobileNetV2\\(8 branches)}} & \multirow{2}[2]{*}{Mocap} \\
    & & & & \\
    \midrule
    \multirow{2}[2]{*}{\textbf{Control error}} & $e_{xy}$ [m] & 0.99 & \textbf{0.68} & 0.18 \\
    & $e_\theta$ [rad] & 0.75 & \textbf{0.49} & 0.21 \\
    \bottomrule
    \end{tabular}
}
\end{table}

\subsection{Onboard security demonstration}

In the supplementary video, we demonstrate our system's security capabilities in an in-field qualitative demonstration.
Initially, the nano-drone performs the ``follow-me'' task in normal conditions without attacks nor security mechanisms in place.
This results in good tracking performance, with the nano-drone able to precisely follow the human subject (never seen in training).
After a few tens of seconds of flight, we simulate an attack in which the fog node continuously sends malicious tensors to the edge.
This malicious tensor encodes a final regression output of $x<\SI{1.5}{\meter}$, which forces the drone to fly away from the subject, i.e., perceived as too close.
Then, the same attack is repeated, but this time having the probabilistic security mechanism active.
In this case, the attack is detected within \SI{1}{\second}, and the edge node reacts by triggering a predefined emergency behavior, i.e., hovering in place.
Once the attack terminates, the fog tensor returns to be trusted by the onboard security mechanism, and the nano-drone resumes following the human subject.
\section{Conclusion} \label{sec:conclusion}

We present a probabilistic security mechanism built on top of an edge-fog system embodied by a resource-constrained nano-drone (edge) and a powerful remote commodity laptop (fog).
By designing a novel MobileNetV2-based CNN, whose heaviest central part is a multi-branch model, we can offload the vast majority of the computation to the fog node while replicating a minimal part on edge to verify its trustworthiness.
We demonstrate our approach's effectiveness, showing \textit{i}) an increased prediction capability of our pose estimation task by a mean +0.19 $R^2$ score compared to a SoA single-branch CNN, and \textit{ii}) enabling our system to detect malicious data infiltration between edge and fog with 95\% probability within \SI{2}{\second} of data exchange.

%
%
\section{Acknowledgments}
This work was partially supported by the Secure Systems Research Center (SSRC) of the UAE Technology Innovation Institute (TII).

%

\bibliographystyle{abbrv}
\bibliography{bibliography}

\begin{thebibliography}{10}

\bibitem{nanoflownet}
R.~J. Bouwmeester et~al.
\newblock {NanoFlowNet}: Real-time dense optical flow on a nano quadcopter.
\newblock {\em arXiv preprint arXiv:2209.06918}, 2022.

\bibitem{ceredahandling}
E.~Cereda et~al.
\newblock Handling pitch variations for visual perception in {MAVs}: Synthetic
  augmentation and state fusion.
\newblock In {\em IMAV}, 2022.

\bibitem{scaramuzza_alphapilot}
P.~Foehn et~al.
\newblock {AlphaPilot}: Autonomous drone racing.
\newblock {\em Autonomous Robots}, 46(1):307--320, 2022.

\bibitem{multibranchadversarial}
T.~Hu et~al.
\newblock Triple wins: Boosting accuracy, robustness and efficiency together by
  enabling input-adaptive inference.
\newblock In {\em ICLR}, 2020.

\bibitem{jung2018perception}
S.~Jung et~al.
\newblock Perception, guidance, and navigation for indoor autonomous drone
  racing using deep learning.
\newblock {\em IEEE Robotics and Automation Letters}, 3(3):2539--2544, 2018.

\bibitem{khanna2020internet}
A.~Khanna et~al.
\newblock Internet of things ({IoT}), applications and challenges: a
  comprehensive review.
\newblock {\em Wireless Personal Communications}, 114:1687--1762, 2020.

\bibitem{9869931}
L.~Lamberti et~al.
\newblock {Tiny-PULP-Dronets}: Squeezing neural networks for faster and lighter
  inference on multi-tasking autonomous nano-drones.
\newblock In {\em 2022 IEEE 4th International Conference on Artificial
  Intelligence Circuits and Systems (AICAS)}, pages 287--290, 2022.

\bibitem{hydranet}
R.~T. Mullapudi et~al.
\newblock {HydraNets}: Specialized dynamic architectures for efficient
  inference.
\newblock In {\em Proceedings of the IEEE Conference on Computer Vision and
  Pattern Recognition (CVPR)}, June 2018.

\bibitem{9654841}
H.~Nguyen et~al.
\newblock Model predictive control for micro aerial vehicles: A survey.
\newblock In {\em 2021 European Control Conference (ECC)}, 2021.

\bibitem{6595932}
D.~Palossi et~al.
\newblock {GPU-SHOT}: Parallel optimization for real-time 3d local description.
\newblock In {\em 2013 IEEE Conference on Computer Vision and Pattern
  Recognition Workshops}, pages 584--591, 2013.

\bibitem{pulp-dronet}
D.~Palossi et~al.
\newblock An open source and open hardware deep learning-powered visual
  navigation engine for autonomous nano-uavs.
\newblock In {\em 2019 15th International Conference on Distributed Computing
  in Sensor Systems (DCOSS)}, pages 604--611. IEEE, 2019.

\bibitem{pulp-frontnet}
D.~Palossi et~al.
\newblock Fully onboard {AI}-powered human-drone pose estimation on
  ultralow-power autonomous flying nano-{UAVs}.
\newblock {\em IEEE Internet of Things Journal}, 9(3):1913--1929, 2021.

\bibitem{deeppilot_2020}
L.~O. {Rojas-Perez} et~al.
\newblock {{DeepPilot}}: {{A CNN}} for {{Autonomous Drone Racing}}.
\newblock {\em Sensors}, 20(16):4524, Jan. 2020.

\bibitem{mobilenetv2}
M.~Sandler et~al.
\newblock {MobileNetV2}: Inverted residuals and linear bottlenecks.
\newblock In {\em Proceedings of the IEEE Conference on Computer Vision and
  Pattern Recognition (CVPR)}, June 2018.

\bibitem{efficientnetv2}
M.~Tan et~al.
\newblock {EfficientNetV2}: Smaller models and faster training.
\newblock In {\em 38th International Conference on Machine Learning (ICML)},
  volume 139, pages 10096--10106, Jul 2021.

\bibitem{branchynet}
S.~Teerapittayanon et~al.
\newblock {BranchyNet}: Fast inference via early exiting from deep neural
  networks.
\newblock In {\em 2016 23rd International Conference on Pattern Recognition
  (ICPR)}, pages 2464--2469, 2016.

\bibitem{deepsubensembles}
M.~Valdenegro-Toro.
\newblock Deep sub-ensembles for fast uncertainty estimation in image
  classification.
\newblock {\em arXiv preprint arXiv:1910.08168}, 2019.

\bibitem{9399464}
M.~Yahuza et~al.
\newblock Internet of drones security and privacy issues: Taxonomy and open
  challenges.
\newblock {\em IEEE Access}, 9:57243--57270, 2021.

\bibitem{2014visualizecnn}
M.~D. Zeiler et~al.
\newblock Visualizing and understanding convolutional networks.
\newblock In {\em Computer Vision -- ECCV 2014}, pages 818--833, 2014.

\bibitem{dronecap}
X.~Zhou et~al.
\newblock Human motion capture using a drone.
\newblock In {\em 2018 IEEE International Conference on Robotics and Automation
  (ICRA)}, 2018.

\end{thebibliography}
\end{document}